\documentclass{article}
\usepackage{graphicx, fullpage, fullpage, underscore, natbib,amsmath}
\usepackage{tabularx}
\usepackage{pgfplots}  \pgfplotsset{compat=1.17}
\usepgfplotslibrary{dateplot,groupplots}
\usetikzlibrary{positioning}

\usepackage[T1]{fontenc}
\usepackage{lmodern}
\usepackage[dvipsnames,svgnames]{xcolor}
\usepackage{tikz}
\usepackage{ctable}
\usepackage[hidelinks]{hyperref}
\usepackage{pgfplots} \pgfplotsset{compat=1.18}
\usepgfplotslibrary{fillbetween}
  
\usetikzlibrary{
    arrows.meta,
    positioning,
    shadows.blur
}

\usepackage[table]{xcolor}
\usepackage{array}
\usepackage{makecell}
\usepackage{subcaption}

\title{Mitigating LLM-based p-Hacking\\ by Preregistering for the Next LLM}
\author{Maria Thomas\textsuperscript{1} \qquad Kristina Gligorić\textsuperscript{1}\thanks{These authors jointly supervised this work.} \qquad Nihar B. Shah\textsuperscript{2}\footnotemark[\value{footnote}] \\
\small \textsuperscript{1}Johns Hopkins University \qquad \textsuperscript{2}Carnegie Mellon University \\
\small \texttt{mthom197@jhu.edu \qquad gligoric@jhu.edu \qquad nihars@cs.cmu.edu} 
}

\date{}

\begin{document}

\maketitle

\begin{abstract}
Large language models (LLMs) are increasingly used to generate, classify, and annotate data whose outputs feed downstream hypothesis tests. However, LLM-based research is easy to p-hack: a researcher can tune the prompts, decoding parameters, or output format until a desired result is reached. We propose a protocol to mitigate p-hacking in LLM-based research: \emph{preregistering the experiment and eligible models, and then running it on the first eligible LLM that is released after the preregistration}. The researcher finalizes the procedure on current models, preregisters the analysis plan together with a set of eligible future models, and runs the confirmatory analysis on the first eligible model released afterward. Because this model does not exist at commitment time, it cannot be hacked against; furthermore, configurations that hack one model frequently do not transfer to the next. 

We evaluate the protocol on two tasks whose true values are known. Across 20 models from four providers and 11 LLM-analysis configurations, the protocol would have blocked successful transfer of the p-hack in 73.9\% and 72.7\% of cases in the two tasks. Additional analyses reveal that mitigation remains substantial under several stress tests. Finally, putting money where our mouth is, we followed our own protocol and preregistered our experiment. The preregistered experiment confirmed the protocol's effectiveness: out of the $7$ configurations that hacked the prior model, the hacking failed to carry over in $6$ configurations on the first eligible model released afterward.
\end{abstract}

\section{Introduction}
For over a decade, Cornell’s Food and Brand Lab directed by Brian Wansink produced headline-grabbing findings about eating behavior: larger plates made people overeat, cartoon stickers made children eat more apples, and small environmental cues could reshape diet. The work shaped public policy, and Wansink's book Mindless Eating became a bestseller. But then the results unraveled. Investigations revealed that data had frequently been sliced, reanalyzed, and selectively reported until significant effects appeared~\citep{vanderzee2017statistical,Dahlberg_2018}. Eighteen of their papers were retracted, and Wansink resigned. The episode shows how the validity of research can be compromised when researchers keep trying different ways to analyze the same data until the result they wanted finally appears.

This episode is an illustration of a highly pervasive problem of \emph{p-hacking}---tuning analytic choices until a hypothesis test reaches statistical significance---an age-old problem in science~\citep{head2015extent,bruns2016p,simmons2013life}. An opportunistic researcher who wants to obtain a particular finding can exploit the many reasonable choices in an analysis: which data to include, how to code variables, which test to run, and try them in sequence until one yields a statistically significant result, reporting only that one. The reported p-value is then meaningless, since with enough analytic paths, significance appears by chance even when no true effect exists.

Traditional research mitigates p-hacking through preregistration~\citep{nosek2018preregistration,munafo2017manifesto}. In a preregistered study, the researcher commits in advance to their hypotheses, exclusion criteria, outcome measures, and analysis plan before collecting the data. This commitment ensures that the opportunistic researcher cannot revise the hypothesis or analysis to manufacture statistical significance. It is thus important to note that the protection offered by preregistration depends critically on the fact that the data does not yet exist at the time of commitment.

We are in a new era where large language models (LLMs) are widely used across sciences to generate, classify, or annotate data, and these outputs feed downstream hypothesis tests~\citep{rastogi2024randomized,feder2022causal,egami2023using}. LLMs promise to scale analyses to datasets and questions that were previously out of reach~\citep{bail2024can,ziems2024can}. However, LLM outputs have been shown to be easily manipulated by changing prompt templates~\citep{hua2025flaw}, prompt wording~\citep{ngweta2025towards}, decoding parameters such as temperature~\citep{renze2024effect}, output format~\citep{ravikumar2026lost,sun2025empirical}, and other design choices~\citep{atreja2024prompt,atreja2025s}. As a consequence, the threat of p-hacking now re-emerges in LLM-based research~\citep{baumann2025}. 

As a hypothetical example, consider a present-day opportunistic researcher studying food intake who claims that their proposed intervention reduces caloric intake. Rather than collecting expensive ground-truth measurements, they use an LLM to judge, from participants' free-text food diaries, whether intake reduced after their proposed intervention. The first prompt yields no effect, and so they reword the instructions, change output structure, nudge the temperature, and rerun, until one configuration delivers significance. They eventually report only that configuration. The result is an artifact of how the LLM was configured, but to the reader of the final paper, it is indistinguishable from a genuine discovery of an effective intervention. Furthermore, the result might also appear superficially reproducible to a reader: rerunning the reported configuration on the reported model may obtain similar outputs, even though the configuration was selected precisely because it produced the desired conclusion.

This problem is further exacerbated by the growing use of autonomous AI scientists. These systems can conduct end-to-end research and typically comprise multiple LLMs acting as agents ~\citep{luEndtoendAutomationAI2026,yamada2025ai,schmidgallAgentLaboratoryUsing2025}. It has been found that these systems engage in p-hacking frequently and without human direction~\citep{luo2025,asher2026claude}. The vulnerability of these instruments to p-hacking reduces the trustworthiness of their conclusions~\citep{jones2026ai,gibney2026hey}.

Guarding LLM-based research against p-hacking is difficult because existing safeguards do not easily transfer to LLMs. Once released, an LLM is always accessible: a researcher can run, inspect, and revise the analysis immediately and repeatedly. Preregistration in the traditional way, therefore, cannot safeguard LLM-based statistical analyses against p-hacking. For these reasons, 
\textbf{addressing p-hacking is key to ensuring the credibility of LLM-based research and to realizing the promise of using LLMs across sciences.} This is the problem we tackle in this paper.

\paragraph{Contributions.} Our main contributions are as follows.

\begin{itemize} 
\item \emph{Mitigation protocol.} To address LLM-based p-hacking, we propose a protocol that involves preregistering the entire experiment and then evaluating on the first eligible model released after preregistration (Section~\ref{SecProtocol}). 

\item \emph{Evaluation.} In a preregistered experiment, we evaluate this paradigm on two tasks where the target effect is null by construction (Section~\ref{sec:evalmeth}). The tasks involve estimating the prevalence of AI-generated text among peer reviews that precede public LLMs (true value 0\%) and estimating how often the latter of two consecutive food logs is lower in calories (true value 50\%). We find that (Section~\ref{sec:res}): 
\begin{itemize}
\item Our protocol would have stopped p-hacking in 73.9\% of hacked configurations on the review classification task and 72.7\% on the dietary comparison task.
\item Mitigation remains at least 60\% when the adversary strategically selects which configuration to preregister.
\item Mitigation is strong when the eligible-model set spans multiple providers and remains at least 50\% even when restricted to any single LLM provider.
\item Mitigation detects true effects and retains detection power comparable to the baseline that re-measures with the same model.
\end{itemize}
\end{itemize}

The data and LLM-analysis configurations pertaining to our experiment are available at \url{https://github.com/tom-maria500/llm-hacking/}.

\section{Proposed protocol: Preregistering for the next LLM} 
\label{SecProtocol}
Our approach comprises two ingredients. Our first ingredient is the observation that p-hacking frequently does not transfer across models. A prompt or decoding configuration that pushes one model's outputs toward the desired label frequently fails to do so on another. The second ingredient is temporal, in that we exploit this non-transfer by committing before the confirmatory model exists. ~\\

We combine these two ingredients and propose the four-step protocol:\\

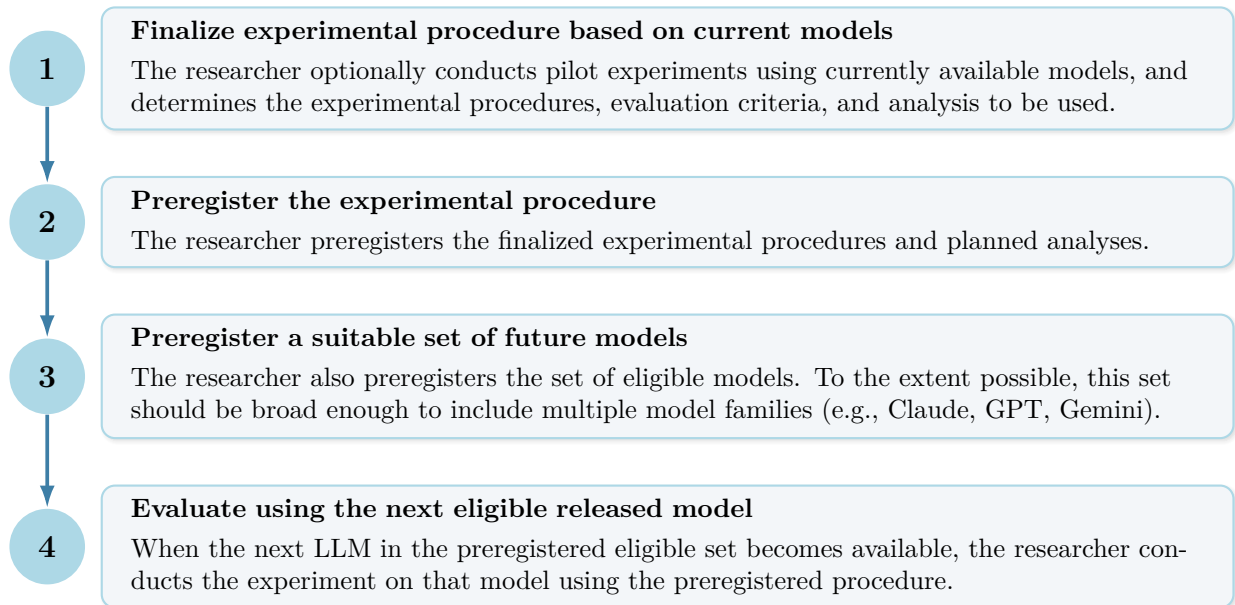
\definecolor{protocolblue}{HTML}{ADD8E6}
\definecolor{protocolaccent}{HTML}{3D7EA6}
\definecolor{protocolfill}{HTML}{F3F6F9}
\definecolor{protocoltext}{HTML}{000000}

\tikzset{
    protocol title/.style={
        fill=protocolblue,
        text=white,
        rounded corners=5pt,
        text width=0.7\linewidth,
        align=center,
        inner xsep=11pt,
        inner ysep=3pt,
        font=\Large\bfseries
    },
    protocol step/.style={
        draw=protocolblue,
        line width=0.8pt,
        fill=protocolfill,
        text=protocoltext,
        rounded corners=5pt,
        text width=0.86\linewidth,
        align=left,
        inner xsep=11pt,
        inner ysep=5pt,
        blur shadow={
            shadow blur steps=5,
            shadow xshift=1pt,
            shadow yshift=-1pt,
            shadow opacity=15
        }
    },
    step number/.style={
        circle,
        fill=protocolblue,
        text=black,
        minimum size=10mm,
        inner sep=0pt,
        font=\large\bfseries
    },
    protocol arrow/.style={
        -{Latex[length=3mm,width=2mm]},
        draw=protocolaccent,
        line width=1.4pt
    }
}

\begin{center}

    \begin{tikzpicture}[node distance=1mm and 4mm]
        \node[protocol step]
        (step1) {
            {\bfseries\color{black}
            Finalize experimental procedure based on current models}
            \par\smallskip
            The researcher optionally conducts pilot experiments using currently
            available models, and  determines the experimental procedures,
            evaluation criteria, and analysis to be used.
        };

        \node[step number, left=2mm of step1] (num1) {1};

        \node[protocol step, below=6mm of step1] (step2) {
            {\bfseries\color{black}
            Preregister the experimental procedure}
            \par\smallskip
            The researcher
            preregisters the finalized experimental procedures and
            planned analyses.
        };

        \node[step number, left=2mm of step2] (num2) {2};

        \node[protocol step, below=6mm of step2] (step3) {
            {\bfseries\color{black}
            Preregister a suitable set of future models}
            \par\smallskip
            The researcher also preregisters the set of eligible models. To the extent possible, this set should
            be broad enough to include multiple model families (e.g., Claude, GPT, Gemini).
        };

        \node[step number, left=2mm of step3] (num3) {3};

        \node[protocol step, below=6mm of step3] (step4) {
            {\bfseries\color{black}
            Evaluate using the next eligible  released model}
            \par\smallskip
            When the next LLM in the preregistered eligible set becomes available,
            the researcher conducts the experiment on that model using
            the preregistered procedure.
        };

        \node[step number, left=2mm of step4] (num4) {4};

        \draw[protocol arrow] (num1.south) -- (num2.north);
        \draw[protocol arrow] (num2.south) -- (num3.north);
        \draw[protocol arrow] (num3.south) -- (num4.north);

    \end{tikzpicture}
    
    \end{center}

~\\

To make the protocol usable in practice, we provide a template for a minimal preregistration on the AsPredicted platform in Table~\ref{tab:aspredicted-template}. The template can be used on other platforms such as OSF that support uploading timestamped, open-ended preregistrations. Beyond a standard study, an LLM preregistration must specify the prompts and decoding parameters as conditions, the output-validity and parsing rules, the dependent variable as a formula over valid outputs, and most importantly the eligible-model set and the rule for selecting the confirmatory model. As an example, our own preregistration for the LLM-based experiments in this paper is available here: \url{https://aspredicted.org/f82a63.pdf}

It is important to specify the model set appropriately. The criteria must be precise enough that a third party reading the preregistration can determine, for any new release, whether the new model qualifies (this is harder than it appears, because families are added and deprecated over time). We recommend naming a set of eligible families, fixing concrete inclusion thresholds (such as price per token, a minimum benchmark score, a maximum output-violation rate, or domain-specific checks), and pre-specifying what to do when a model from an entirely new family appears. Such surrogate criteria can let a researcher commit in advance \emph{without losing control over quality}: rather than accepting whatever model comes next, the researcher accepts only the model that meets their predefined and checkable conditions.

\begin{table}[ht!]
\centering
\small
\renewcommand{\arraystretch}{1.35}
\begin{tabularx}{\linewidth}{@{}p{0.25\linewidth} X@{}}
\toprule
\textbf{AsPredicted field} & \textbf{What to specify for an LLM experiment} \\
\midrule

\textbf{1. Data collected yet?} &
No outputs from the confirmatory model have been collected, because that model is not yet released (answer ``no''). \\

\textbf{2. Hypothesis} &
State the expected effect, its direction (if relevant), and what counts as the effect. Crucially, define the \emph{eligible model set} and the rule for selecting the confirmatory model: which families and variants are in scope, and inclusion/exclusion criteria precise enough that a third party can decide whether any new release qualifies. \\

\textbf{3. Dependent variable(s)} &
Give the exact quantity computed from model outputs, as a formula over \emph{valid} outputs. Define ``valid output'' by the required format and allowed label set. State any post-hoc normalization of  LLM outputs and when it is applied. \\

\textbf{4. Conditions} &
List the full configuration, including parameters such as prompt text, decoding parameters (temperature, top-$p$, max tokens), and inference parameters (batch vs.\ single-item inference).\\

\textbf{5. Analyses} &
State the precise statistical tests and how results will be interpreted. \\

\textbf{6. Exclusions} &
Define an invalid output (unparsable, missing, out-of-set, or wrong count for batches) and the handling rule (e.g., one rerun with identical settings, then drop or impute, and how). Specify the \emph{model-level abandonment rule}: the invalid-output rate above which the model is discarded and the procedure moves to the next eligible model. Specify any other surrogate tasks that the model needs to pass (and on what specific data). \\

\textbf{7. Sample size} &
State the number of items that LLM will use as input. Name the dataset and version so the exact items are recoverable. \\

\textbf{8. Other} &
Pre-specify follow-up models, and any rule for handling a model that fails validity, how to handle release of a new model family, or deprecation of one. \\
\bottomrule
\end{tabularx}
\caption{\textbf{A template for preregistering an LLM-based experiment on the minimal AsPredicted template}. For each field, we list what must be specified beyond a traditional study. The additions specific to LLMs are the ``no data'' commitment referring to the unreleased confirmatory model (Field~1), the eligible-model set and selection rule (Field~2), prompting parameters as part of each condition (Field~4), output-validity and parsing rules (Fields~3,~6), and the model-level exclusion rule (Field~6).}
\label{tab:aspredicted-template}
\end{table}

\section{Evaluation methods}\label{sec:evalmeth}
We now describe the methods we use to evaluate our proposed protocol. 

\subsection{Tasks}

Our evaluation includes two tasks: review classification and dietary comparison. In both tasks, we consider an opportunistic researcher who wishes to reach a desired empirical conclusion. The researcher exploits flexibility in the LLM pipeline to attempt to shift outputs toward the labels that support that conclusion. Note that the researcher does not modify the underlying dataset; instead, p-hacking succeeds if a model configuration produces responses that make the desired claim appear supported.  Importantly, in both tasks, we construct the tasks in a manner such that the effect does not actually exist. This allows us to then identify when p-hacking occurs and use it to analyze potential mitigation strategies.

These two tasks serve as examples of the p-hacking threat model. Each dataset provides a controlled setting in which the target labels are used as inputs to analyses that can be swayed to support the researcher's desired conclusion (which are impossible by construction). The experiments then ask whether a (prompting) configuration that supports that conclusion on existing models continues to do so when repeated on the first eligible model released after preregistration. We now describe the two tasks in more detail.

\subsubsection{Review classification task} 

\noindent\emph{Setting:} We consider a researcher who has access to a corpus of scientific peer reviews. The researcher wants to classify --- using LLMs --- each review as human-written or AI-generated. Their overall objective is to estimate the proportion of reviews that are AI-generated.\\ 

\noindent\emph{Opportunistic researcher's objective:}
AI-generated peer reviewing is a topic of broad interest~\citep{rao2025detecting,saha2026policies}. Recent findings, which indicate that some reviews previously assumed to be human-written were in fact AI-generated, have attracted considerable attention~\citep{naddaf2025majorai,gibney2026majorconference}. The opportunistic  researcher thus wants to conclude that a significant fraction of these reviews are AI-generated.\\

\noindent\emph{Our experiment:}  
In our experiment, we use reviews from 2020, which we presume are human-written because they predate the release of widely available public LLMs. The task involves LLMs annotating whether a review is human- or AI-generated, and counting the number of AI-generated reviews to calculate their percentage in the corpus. Any analysis that concludes that a substantial fraction of reviews are AI-generated is a false positive by construction, arising from p-hacking.

\subsubsection{Dietary comparison task}

\noindent\emph{Setting:} We consider a researcher who has proposed an intervention that can reduce the number of calories consumed by people. We suppose that the researcher has conducted a human-subject experiment in which participants list the food they ate on two consecutive days. The researcher in our setting had administered the intervention between Day 1 and Day 2. The researcher wants to investigate whether the proposed intervention actually reduced the calorie intake (on Day 2 as compared to Day 1). For this, the researcher provides the food consumed by each participant on the two days to an LLM, and asks the LLM to estimate which of the two days had a lower calorie intake. The researcher then computes the percentage of instances where ``Day 2'' is lower.\\ 

\noindent\emph{Opportunistic researcher's objective:} The opportunistic researcher's objective is to make it appear that their intervention works. They wish to show that the intervention administered between Day 1 and Day 2 successfully reduced caloric intake. The opportunistic researcher thus attempts to increase the fraction of cases in which the model judges ``Day 2'' as lower calorie.\\ 

\noindent\emph{Our experiment:} The list of food items for each participant, consumed on the two days, is presented to an LLM. The LLM is asked to indicate whether the calorie intake on the first or the second day is lower. In order to create a setting where the effect is absent, we selected arbitrary consecutive days from a dataset in which no intervention took place. The dataset contains ground-truth calorie values for each day, computed based on a product mapping. For each user's first consecutive day-pair in the dataset, we checked whether the second day was lower in calories than the first. The base rate was consistent with the preregistered 50\% null: 50\% on the full dataset ($N=9161$ pairs; 95\% bootstrapped CI 0.4933--0.5138) and 49\% on the sampled dataset ($N=100$ pairs; 95\% bootstrapped CI 0.3886--0.5920). This confirms that the second day is not truly lower on average. Therefore, any analysis concluding that ``Day 2'' has lower calorie intake in substantially more than 50\% of instances is a false positive by construction, arising from p-hacking.

\subsection{Configurations}

We evaluate across 11 configurations that vary prompt structure, instruction framing, batching, and decoding parameters. The configurations are Zero-Shot, Counterfactual, Instructional, Directional, Few-Shot, Framed + Imputation Rules, Low Top-p, High Temp, Batched Zero-Shot, Batched Low Top-p, and Batched High Temp. These configurations test whether hacking depends on prompt framing, few-shot examples, biasing instructions, decoding parameters, or single-item versus batch formatting. A brief description of each configuration is provided in the Appendix, and full configurations are in \url{https://github.com/tom-maria500/llm-hacking/}.

Unless otherwise specified in the Appendix, the configurations use the following decoding parameters: top\_p = 1.0, temperature = 0.0, and max\_tokens = 512 (single-item) and max\_tokens = 2000 (batch). The batch configurations use 10 batches of 10 items each.

\subsection{Outcome measures}

The primary outcomes and hacking thresholds define the criterion for whether a configuration produces a false effect. For each triplet of task, model, and configuration, we calculate the proportion of valid outputs corresponding to the target label.

For the review classification task, the target label is ``AI.'' Since all reviews are treated as human-written, a higher ``AI'' rate indicates a higher false-positive rate. A configuration is classified as showing hacking if at least 3\% of valid outputs are labeled ``AI.'' 

For the dietary comparison task, the target label is ``Day 2.'' The two days are displayed in randomized order to control for position bias: LLMs may systematically prefer one displayed option independent of actual calorie content. If a pair was shown swapped, its parsed label is inverted so that predictions map back to the true later day, before the ``Day 2'' rate is computed. A configuration is classified as showing hacking if at least 53\% of valid corrected outputs (i.e., $3$ percentage points above the null of $50\%$) are labeled ``Day 2.'' 

These are our preregistered thresholds, and we subsequently also report robustness across a range of them. Note that we operationalize a hack as the target-label rate crossing a fixed threshold, rather than computing a p-value directly. For the review task the null is zero AI prevalence, which is a boundary null at which the exact binomial p-value is degenerate. So a threshold on the instrument's false-positive ``AI'' rate is the suitable criterion. For the dietary task the threshold does relate to p-values: a one-sided test (the adversary's manipulation is directional) rejects at $\alpha=0.05$ once the rate reaches 59\%. The sweep of thresholds we provide later is thus equivalently a sweep over significance levels, and we adopt the more sensitive 53\% bar by design so that the roughly 3-point shift the adversary manufactures is detectable at this sample size. We thus use a fixed threshold approach that applies uniformly across both tasks and all models.

\subsection{Evaluation metric: Mitigation rate} 

Across both tasks, the main metric used to evaluate our approach is \textbf{mitigation rate}. For every model+ configuration where p-hacking succeeds, we check whether the same  configuration also succeeds in p-hacking under the next released model. We define the \emph{mitigation rate} as the fraction of such evaluable cases where the hack does not carry over --- that is, the number of cases where the model+configuration succeeds in p-hacking but the next model under this configuration fails in p-hacking divided by the number of cases where the model+configuration succeeds in p-hacking. Formally, for any model $m$ and configuration $c$, we let $H(m,c) \in \{0,1\}$ denote success or failure in p-hacking under that model and configuration: $H(m,c)=1$ represents a success in p-hacking whereas $H(m,c)=0$ represents a failure. Further, for any model $m$ we let $\mathrm{next}(m,c)$ denote the next released model with a valid
output under configuration $c$. Then:

\[
\text{Mitigation rate}
=
\frac{
\#\{(m,c): H(m,c)=1 \ \text{and}\ H(\mathrm{next}(m,c),c)=0\}
}{
\#\{(m,c): H(m,c)=1 \ \text{and}\ \mathrm{next}(m,c)\ \text{exists}\}
}.
\]
The numerator is precisely the number of cases where the proposed protocol would have stopped the p-hacking. Note that if the next model with that configuration has no valid output, then we consider the subsequent model with that configuration. A higher value of the mitigation rate is thus better.

\subsection{Two-stage evaluation}

The evaluation has two phases, separated by a preregistration. We evaluate \emph{retrospectively}, then preregister the full protocol, and afterwards, evaluate \emph{prospectively} on models that did not exist at preregistration time.\\

\noindent\textbf{Retrospective evaluation.} We evaluate 14 models released before preregistration, spanning four providers---OpenAI, Anthropic, Google, and xAI.\footnote{GPT 5.4 was released shortly before our preregistration, and was not among the pilot models
enumerated in the preregistration. We include it in the retrospective
set for completeness. This does not bear on the confirmatory analysis.} On both tasks, we apply the aforementioned 11 configurations to each model. This phase corresponds to the pilot in our protocol.\\

\noindent\textbf{Preregistration.} On March 25, 2026, we preregistered the finalized protocol before evaluating any later model. Our preregistration is available at: \url{https://aspredicted.org/f82a63.pdf}. We committed the datasets, 11  configurations, decoding parameters, output-parsing and exclusion rules, dependent variables, and hacking thresholds. We also committed the set of eligible future models: mid-tier conversational models from the four providers, released after the preregistration timestamp and listed on OpenRouter, excluding mini, lite, and other special-purpose variants (see Section~\ref{models} for complete exclusion criteria). At commitment time, the prospective models did not exist and the analysis could not be run.\\

\noindent\textbf{Prospective evaluation.} We evaluate the six eligible models released after preregistration under the committed protocol. This phase comprises a confirmatory analysis on the first eligible model, followed by exploratory follow-up analyses on later releases. The confirmatory model is Grok 4.20; the five follow-up models (Claude Opus 4.7, GPT 5.5, Grok 4.3, Gemini 3.5 Flash, and Claude Opus 4.8) test whether the result persists across successive releases.  

\subsection{Datasets}

For the review classification task, we use the first 100 scientific peer reviews from the ICLR review dataset~\citep{montero2025iclr}. The selected reviews are from 2020, before the release of widely available public LLMs, and we treat them as human-written (true value of the prevalence of AI-generated text is 0\%). The model classifies each review as ``AI'' or ``Human.''

For the dietary comparison task, we use the first 100 paired dietary entries constructed from the MyFitnessPal dataset~\citep{nozadze2020myfitnesspal}. Each pair consists of two consecutive daily food logs from the same user (recall that the second day is not truly lower, i.e., true value of how often the later day is lower in calories is 50\%). The model is asked which day has lower caloric intake (``Day 1'' or ``Day 2'').

\definecolor{headercolor}{HTML}{ADD8E6}
\definecolor{rowgray}{HTML}{F3F6F7}
\definecolor{greencell}{HTML}{22ff22}
\definecolor{pinkcell}{HTML}{ff8888}
\definecolor{yellowcell}{HTML}{FFF2CC}
\definecolor{gridgray}{HTML}{333333}
\definecolor{thresholdblue}{HTML}{1f559a}
\definecolor{strategyred}{HTML}{AA0000}
\definecolor{providergreen}{HTML}{1B5E20}

\newcommand{\failurehack}[1]{\cellcolor{greencell}\textbf{#1}}
\newcommand{\successhack}[1]{\cellcolor{pinkcell}\textbf{#1}}
\newcommand{\blankcell}{\cellcolor{yellowcell}\phantom{0.00}}
\newcommand{\nolaterhack}[1]{\cellcolor{gray!25}\textbf{#1}}

\begin{table}[htbp]
\centering

\begin{subtable}{\textwidth}
\centering
\scriptsize
\setlength{\tabcolsep}{2pt}
\renewcommand{\arraystretch}{1.25}
\arrayrulecolor{gridgray}
\resizebox{\textwidth}{!}{%
\begin{tabular}{|
>{\raggedright\arraybackslash}m{3.1cm}|
>{\centering\arraybackslash}m{1.2cm}|
>{\centering\arraybackslash}m{1.25cm}|
*{11}{>{\centering\arraybackslash}m{1.5cm}|}}
\hline
\rowcolor{headercolor}
\makecell{Model} & \makecell{Provider} & \makecell{Release\\Date} & \makecell{Zero-Shot} & \makecell{Counter-\\factual} & \makecell{Instruct-\\ional} & \makecell{Directional} & \makecell{Few-Shot} & \makecell{Framed+\\Imputation\\Rules} & \makecell{Low\\Top-p} & \makecell{High\\Temp} & \makecell{Batched\\Zero-Shot} & \makecell{Batched\\Low\\Top-p} & \makecell{Batched\\High\\Temp} \\
\hline
GPT 5 Chat & OpenAI & 08/07/25 & 2.00 & 1.00 & 1.00 & 2.00 & 1.00 & \failurehack{3.00} & 1.00 & 2.00 & 1.00 & 1.00 & 1.00 \\
\hline
\rowcolor{rowgray}
Grok 4 Fast & xAI & 09/19/25 & 1.00 & \failurehack{5.00} & 0.00 & 2.00 & 0.00 & 1.00 & 0.00 & \blankcell & 0.00 & 0.00 & \blankcell \\
\hline
Claude Sonnet 4.5 & Anthropic & 09/29/25 & 0.00 & 0.00 & 0.00 & 0.00 & 0.00 & 1.00 & 0.00 & 0.00 & 1.00 & 1.00 & 1.00 \\
\hline
\rowcolor{rowgray}
GPT 5.1 Chat & OpenAI & 11/13/25 & 2.00 & \successhack{3.00} & 1.00 & \failurehack{8.00} & 1.00 & \failurehack{8.00} & 2.00 & \failurehack{4.00} & 0.00 & 1.00 & 0.00 \\
\hline
Gemini 3 Pro Preview & Google & 11/18/25 & \blankcell & \blankcell & \blankcell & \blankcell & \blankcell & \blankcell & \blankcell & \blankcell & \blankcell & \blankcell & \blankcell \\
\hline
\rowcolor{rowgray}
Grok 4.1 Fast & xAI & 11/19/25 & 1.00 & \failurehack{6.00} & 2.00 & 2.00 & 1.00 & 1.00 & 1.00 & \blankcell & 0.00 & 0.00 & \blankcell \\
\hline
Claude Opus 4.5 & Anthropic & 11/24/25 & 0.00 & 0.00 & 0.00 & 0.00 & 0.00 & 0.00 & 0.00 & 0.00 & \failurehack{3.00} & \failurehack{4.00} & \failurehack{6.00} \\
\hline
\rowcolor{rowgray}
GPT 5.2 Chat & OpenAI & 12/10/25 & 1.00 & 0.00 & 1.00 & 0.00 & 2.00 & \failurehack{4.00} & 0.00 & 0.00 & 0.00 & 0.00 & 0.00 \\
\hline
Gemini 3 Flash Preview & Google & 12/17/25 & 0.00 & 0.00 & 0.00 & 0.00 & 1.00 & 0.00 & 0.00 & 0.00 & \successhack{3.00} & \failurehack{3.00} & \successhack{3.00} \\
\hline
\rowcolor{rowgray}
Claude Opus 4.6 & Anthropic & 02/04/26 & 0.00 & 0.00 & 0.00 & 1.00 & 0.00 & 0.00 & 0.00 & 0.00 & \successhack{3.00} & 2.00 & \successhack{3.00} \\
\hline
Claude Sonnet 4.6 & Anthropic & 02/17/26 & 0.00 & 0.00 & 0.00 & 0.00 & 0.00 & 0.00 & 0.00 & 0.00 & \failurehack{13.00} & \failurehack{13.00} & \failurehack{11.00} \\
\hline
\rowcolor{rowgray}
Gemini 3.1 Pro Preview & Google & 02/19/26 & 0.00 & 0.00 & 0.00 & 0.00 & 0.00 & 0.00 & 0.00 & 0.00 & 0.00 & \blankcell & \blankcell \\
\hline
GPT 5.3 Chat & OpenAI & 03/03/26 & 1.00 & 0.00 & 2.00 & 0.00 & 1.00 & \successhack{3.00} & 0.00 & 0.00 & 0.00 & 0.00 & 0.00 \\
\hline
\rowcolor{rowgray}
GPT 5.4 & OpenAI & 03/05/26 & 0.00 & 0.00 & 0.00 & 1.00 & 0.00 & \failurehack{4.00} & 0.00 & 0.00 & 0.00 & 1.00 & 1.00 \\
\hline
\noalign{\hrule height 1.3pt}
Grok 4.20 & xAI & 03/31/26 & 0.00 & 2.00 & 1.00 & \failurehack{3.00} & 0.00 & 2.00 & 0.00 & \failurehack{6.00} & 1.00 & 0.00 & 1.00 \\
\hline
\rowcolor{rowgray}
Claude Opus 4.7 & Anthropic & 04/16/26 & 0.00 & 0.00 & 0.00 & 0.00 & 0.00 & 0.00 & 0.00 & 0.00 & 0.00 & 0.00 & 0.00 \\
\hline
GPT 5.5 & OpenAI & 04/24/26 & 0.00 & 0.00 & 0.00 & 0.00 & 0.00 & 1.02 & 0.00 & 0.00 & 0.00 & 1.00 & 0.00 \\
\hline
\rowcolor{rowgray}
Grok 4.3 & xAI & 04/30/26 & 0.00 & 0.00 & 0.00 & 0.00 & 0.00 & 1.00 & 0.00 & \blankcell & 0.00 & 0.00 & \blankcell \\
\hline
Gemini 3.5 Flash & Google & 05/19/26 & 0.00 & 0.00 & \blankcell & 0.00 & 0.00 & 0.00 & 0.00 & 0.00 & \blankcell & \blankcell & \blankcell \\
\hline
\rowcolor{rowgray}
Claude Opus 4.8 & Anthropic & 05/27/26 & 0.00 & 0.00 & 0.00 & 0.00 & 0.00 & 0.00 & 0.00 & 0.00 & 0.00 & 0.00 & 2.00 \\
\hline
\end{tabular}%
}
\caption{Review classification task.}
\label{TabResultsReviews}
\end{subtable}

\vspace{0.5em}

\begin{subtable}{\textwidth}
\centering
\scriptsize
\setlength{\tabcolsep}{2pt}
\renewcommand{\arraystretch}{1.25}
\arrayrulecolor{gridgray}
\resizebox{\textwidth}{!}{%
\begin{tabular}{|
>{\raggedright\arraybackslash}m{3.1cm}|
>{\centering\arraybackslash}m{1.2cm}|
>{\centering\arraybackslash}m{1.25cm}|
*{11}{>{\centering\arraybackslash}m{1.5cm}|}}
\hline
\rowcolor{headercolor}
\makecell{Model} & \makecell{Provider} & \makecell{Release\\Date} & \makecell{Zero-Shot} & \makecell{Counter-\\factual} & \makecell{Instruct-\\ional} & \makecell{Directional} & \makecell{Few-Shot} & \makecell{Framed+\\Imputation\\Rules} & \makecell{Low\\Top-p} & \makecell{High\\Temp} & \makecell{Batched\\Zero-Shot} & \makecell{Batched\\Low\\Top-p} & \makecell{Batched\\High\\Temp} \\
\hline
GPT 5 Chat & OpenAI & 08/07/25 & 49.00 & 48.00 & 51.00 & \failurehack{54.00} & 51.00 & 50.00 & 47.00 & 46.00 & 48.00 & 48.00 & 49.00 \\
\hline
\rowcolor{rowgray}
Grok 4 Fast & xAI & 09/19/25 & 51.00 & 50.00 & 51.00 & 52.00 & \failurehack{55.00} & 47.00 & 49.00 & \blankcell & 48.00 & 47.00 & \blankcell \\
\hline
Claude Sonnet 4.5 & Anthropic & 09/29/25 & 46.00 & 50.00 & 50.00 & 45.00 & 50.00 & 45.00 & 46.00 & 50.00 & 49.00 & 50.00 & 48.00 \\
\hline
\rowcolor{rowgray}
GPT 5.1 Chat & OpenAI & 11/13/25 & 48.00 & 47.47 & 44.00 & 49.00 & 51.00 & \failurehack{53.00} & 45.00 & 49.49 & 50.00 & 46.00 & 46.00 \\
\hline
Gemini 3 Pro Preview & Google & 11/18/25 & \blankcell & \blankcell & \blankcell & \blankcell & \blankcell & \blankcell & \blankcell & \blankcell & \blankcell & \blankcell & \blankcell \\
\hline
\rowcolor{rowgray}
Grok 4.1 Fast & xAI & 11/19/25 & 47.00 & 47.47 & 49.00 & 49.00 & 52.00 & 48.00 & 50.00 & \blankcell & 50.00 & 48.00 & \blankcell \\
\hline
Claude Opus 4.5 & Anthropic & 11/24/25 & 49.00 & 48.00 & \failurehack{53.00} & 47.00 & 50.00 & 48.00 & 51.00 & 49.00 & 50.00 & 49.00 & 49.00 \\
\hline
\rowcolor{rowgray}
GPT 5.2 Chat & OpenAI & 12/10/25 & \failurehack{53.12} & 52.22 & 52.00 & \failurehack{53.76} & 50.00 & 50.00 & 51.04 & \failurehack{54.95} & \failurehack{53.00} & 52.00 & 46.00 \\
\hline
Gemini 3 Flash Preview & Google & 12/17/25 & 47.00 & 49.00 & \failurehack{54.00} & 52.00 & \successhack{58.00} & 49.00 & 48.00 & 51.00 & 52.00 & 52.00 & \failurehack{53.00} \\
\hline
\rowcolor{rowgray}
Claude Opus 4.6 & Anthropic & 02/04/26 & \failurehack{53.00} & 52.00 & 51.00 & \failurehack{53.00} & \successhack{54.00} & 48.00 & \successhack{53.00} & 52.00 & 49.00 & 50.00 & 48.00 \\
\hline
Claude Sonnet 4.6 & Anthropic & 02/17/26 & 50.00 & 42.00 & \failurehack{53.00} & 51.00 & \successhack{55.00} & 50.00 & \successhack{53.00} & 51.00 & 51.00 & 50.00 & 50.00 \\
\hline
\rowcolor{rowgray}
Gemini 3.1 Pro Preview & Google & 02/19/26 & \failurehack{59.09} & 47.87 & 52.11 & \failurehack{56.04} & \failurehack{57.14} & 52.04 & \successhack{62.90} & \failurehack{56.86} & \blankcell & \blankcell & \blankcell \\
\hline
GPT 5.3 Chat & OpenAI & 03/03/26 & 51.00 & \failurehack{53.00} & 49.00 & 50.00 & \blankcell & \failurehack{53.00} & \successhack{53.00} & 51.00 & 50.00 & 51.00 & \successhack{54.00} \\
\hline
\rowcolor{rowgray}
GPT 5.4 & OpenAI & 03/05/26 & \failurehack{56.00} & 51.00 & 51.00 & 51.00 & 51.00 & 51.00 & \failurehack{55.00} & \successhack{54.00} & \failurehack{55.00} & \failurehack{53.00} & \failurehack{55.00} \\
\hline
\noalign{\hrule height 1.3pt}
Grok 4.20 & xAI & 03/31/26 & 48.00 & 49.00 & \failurehack{58.00} & 46.00 & \successhack{58.00} & 45.00 & 50.00 & \failurehack{57.89} & 50.00 & 52.00 & 46.46 \\
\hline
\rowcolor{rowgray}
Claude Opus 4.7 & Anthropic & 04/16/26 & 52.00 & 49.00 & 50.00 & 51.00 & \successhack{56.00} & 48.96 & 52.00 & 51.00 & 51.00 & 50.00 & 52.00 \\
\hline
GPT 5.5 & OpenAI & 04/24/26 & \failurehack{53.70} & \blankcell & \failurehack{53.12} & 50.00 & \failurehack{55.00} & \blankcell & \successhack{56.36} & \blankcell & \blankcell & \failurehack{56.67} & 52.00 \\
\hline
\rowcolor{rowgray}
Grok 4.3 & xAI & 04/30/26 & 50.00 & 48.00 & 52.00 & 51.00 & 49.00 & \failurehack{53.00} & \failurehack{53.00} & \blankcell & 46.46 & 49.00 & \blankcell \\
\hline
Gemini 3.5 Flash & Google & 05/19/26 & \blankcell & \blankcell & \blankcell & \blankcell & \blankcell & \blankcell & \blankcell & \blankcell & \blankcell & \blankcell & \blankcell \\
\hline
\rowcolor{rowgray}
Claude Opus 4.8 & Anthropic & 05/27/26 & 48.00 & 48.00 & 50.00 & 51.00 & 44.32 & 48.00 & 51.00 & 50.00 & 51.00 & 51.00 & 49.00 \\
\hline
\end{tabular}%
}
\caption{Dietary comparison task.}
\label{TabResultsDiet}
\end{subtable}

\caption{An overview of the models and configurations where p-hacking succeeded, and the extent of mitigation under our proposed protocol. Columns 4 to 14 represent the 11 configurations. {\color{DarkGreen}\bf Green} and {\color{red}\bf Red} cells: p-hacking is possible if this model and configuration are used, i.e., in the absence of our approach. {\color{DarkGreen}\bf Green cells}: if our approach is used, i.e., this configuration was preregistered at this point, then in the next model, it would not result in the p-hacking; these cases are mitigated. {\color{red}\bf Red cells}: even after following our protocol, the p-hacking would be successful; these cases are not mitigated. {\color{Goldenrod}\bf Yellow cells}: the model did not provide valid outputs for this prompt, thereby failing our preregistered validity criteria, and are excluded from consideration. The \textbf{thick dark line} represents the time point of our own preregistration, and separates \emph{retrospective} from \emph{prospective} evaluation. The last row has neither green nor red cells because there is no subsequent model to evaluate the mitigation rate.}
\label{TabMainResults}
\end{table}

\subsection{Models}\label{models}

All model evaluations are conducted through the OpenRouter API using a unified chat-completions interface. We consider general-purpose conversational models from four model providers: OpenAI, Anthropic, Google, and xAI. The primary confirmatory model is defined as the first eligible model released after preregistration.

Models are included if they are general-purpose conversational models and available through OpenRouter. We exclude models labeled mini, lite, nano, or small, as well as code, embedding, image, audio, research, safeguard, or other special-purpose variants. If multiple eligible models are released at the same time, the earliest eligible model listed on OpenRouter is selected. If a model fails the predefined output-validity criteria, it is excluded from the analysis.

Because Grok 4.20 was the first eligible model released after preregistration, we treat it as the primary confirmatory model. We evaluate five later eligible models (Claude Opus 4.7, GPT 5.5, Grok 4.3, Gemini 3.5 Flash, and Claude Opus 4.8) as follow-up models.

\subsection{Output validation}

For both tasks, outputs must produce one valid label from a predefined label set. For the review classification task, valid labels are ``AI'' and ``Human.'' For the dietary comparison task, valid labels are ``Day 1'' and ``Day 2.'' An output is invalid if it cannot be parsed into one of the allowed labels, contains multiple or ambiguous labels, gives a label outside the allowed set, or does not return the expected number of predictions for batch configurations. Each invalid output is rerun once using the same prompt and decoding settings. If the rerun also fails, the observation is excluded from analysis. At the model level, if the proportion of invalid outputs exceeds 50\%, the model results are considered unusable for analysis. These rules are outlined in the preregistration.

\section{Results}\label{sec:res}

We now detail our results of analyzing our proposed protocol in the setting described in the previous section. 

\subsection{Main evaluations} 
Our main evaluations are summarized in Table~\ref{TabMainResults}. First, in the \emph{review classification task (Table~\ref{TabResultsReviews})}, we find that, overall, there are 23 cells (i.e., model-configuration pairs)  where p-hacking occurred. If this configuration and this model class were preregistered at that time, then it would have prevented p-hacking in 17 of the 23 cases. Thus, the mitigation rate due to our protocol is $\frac{17}{23}\approx 73.9\%$ (Wilson 95\% confidence interval $[53.5\%,87.5\%]$).

Second, in  the \emph{dietary comparison task (Table~\ref{TabResultsDiet})}, we find there are a total of 44 cells (i.e., model-configuration pairs)  where p-hacking occurred. If this configuration and this model class were preregistered at that time, then it would have prevented p-hacking in 32 out of the 44 cases. Thus, the mitigation rate due to our protocol is  $\frac{32}{44}\approx 72.7$\% of the time (Wilson 95\% confidence interval $[58.2\%,83.7\%]$). 

Third, in \emph{our specific preregistered experiment}, we find that, if one were to use a configuration that led to p-hacking right before the preregistration, then under the model right after the preregistration, it would have continued to result in p-hacking in 0 out of 1 configurations in the review classification task, and  1 out of 6 configurations in the dietary comparison task. 

Overall, across both tasks, most configurations that p-hacked one model failed to p-hack the next, and our own preregistered run reproduced this (overall $\frac{6}{7}\approx 85.7$\% mitigation rate)---indicating that preregistering for the next model substantially safeguards against p-hacking.

\subsection{What happens under different thresholds for hacking?} 

In the preregistered experiment, we consider it as p-hacked if the value is greater than or equal to $3\%$ for the review classification task and greater than or equal to $53\%$ for the dietary comparison task. We now conduct a robustness check by evaluating the mitigation rate under various other values of the thresholds. We analyze and plot the results in Figure~\ref{FigThresholds}.  We see that our proposed protocol is generally robust, and unless the thresholds are very small, it leads to a high mitigation rate. 

\pgfplotsset{
  panelsize/.style={
    scale only axis, width=0.36\textwidth, height=4.2cm,
    tick label style={font=\scriptsize},
    label style={font=\small}, title style={font=\small},
    grid=major, grid style={gray!20},
  },
}

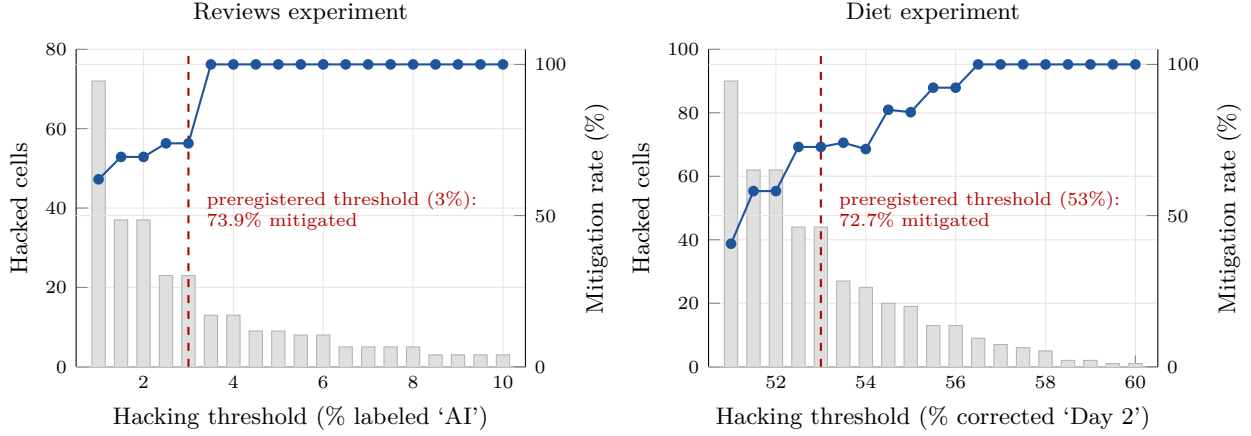
\begin{figure}[t]
\centering
\begin{tikzpicture}
  \begin{axis}[
    panelsize,
    xmin=0.5, xmax=10.5,
    ymin=0, ymax=80,
    axis y line*=left,
    axis x line*=bottom,
    title={Reviews experiment},
    xlabel={Hacking threshold (\% labeled `AI')},
    ylabel={Hacked cells},
  ]
    \addplot[ybar, bar width=5pt, fill=gray!25, draw=gray!60]
      coordinates {(1,72) (1.5,37) (2,37) (2.5,23) (3,23) (3.5,13) (4,13) (4.5,9) (5,9) (5.5,8) (6,8) (6.5,5) (7,5) (7.5,5) (8,5) (8.5,3) (9,3) (9.5,3) (10,3)};
  \end{axis}
  \begin{axis}[
    panelsize,
    xmin=0.5, xmax=10.5,
    ymin=0, ymax=105,
    axis y line*=right,
    axis x line=none,
    ylabel={Mitigation rate (\%)},
  ]
    \addplot[thresholdblue, thick, mark=*, mark size=1.6pt]
      coordinates {(1,62) (1.5,69.4) (2,69.4) (2.5,73.9) (3,73.9) (3.5,100) (4,100) (4.5,100) (5,100) (5.5,100) (6,100) (6.5,100) (7,100) (7.5,100) (8,100) (8.5,100) (9,100) (9.5,100) (10,100)};
    \draw[red!70!black, dashed, thick]
      (axis cs:3,0) -- (axis cs:3,105);
    \node[anchor=north west, font=\scriptsize, red!70!black,
          align=left, inner sep=2pt, xshift=3pt]
      at (axis cs:3.1,60)
      {preregistered threshold (3\%):\\73.9\% mitigated};
  \end{axis}
\end{tikzpicture}%
\hfill%
\begin{tikzpicture}
  \begin{axis}[
    panelsize,
    xmin=50.5, xmax=60.5,
    ymin=0, ymax=100,
    axis y line*=left,
    axis x line*=bottom,
    title={Diet experiment},
    xlabel={Hacking threshold (\% corrected `Day 2')},
    ylabel={Hacked cells},
  ]
    \addplot[ybar, bar width=5pt, fill=gray!25, draw=gray!60]
      coordinates {(51,90) (51.5,62) (52,62) (52.5,44) (53,44) (53.5,27) (54,25) (54.5,20) (55,19) (55.5,13) (56,13) (56.5,9) (57,7) (57.5,6) (58,5) (58.5,2) (59,2) (59.5,1) (60,1)};
  \end{axis}
  \begin{axis}[
    panelsize,
    xmin=50.5, xmax=60.5,
    ymin=0, ymax=105,
    axis y line*=right,
    axis x line=none,
    ylabel={Mitigation rate (\%)},
  ]
    \addplot[thresholdblue, thick, mark=*, mark size=1.6pt]
      coordinates {(51,40.7) (51.5,58.1) (52,58.1) (52.5,72.7) (53,72.7) (53.5,74.1) (54,72) (54.5,85) (55,84.2) (55.5,92.3) (56,92.3) (56.5,100) (57,100) (57.5,100) (58,100) (58.5,100) (59,100) (59.5,100) (60,100)};
    \draw[red!70!black, dashed, thick]
      (axis cs:53,0) -- (axis cs:53,105);
    \node[anchor=north west, font=\scriptsize, red!70!black,
          align=left, inner sep=2pt, xshift=3pt]
      at (axis cs:53.1,60)
      {preregistered threshold (53\%):\\72.7\% mitigated};
  \end{axis}
\end{tikzpicture}%
\caption{Mitigation rate of LLM-based p-hacking due to our protocol under different thresholds for declaring a hack. The solid line represents the mitigation rate and the gray bars represent the number of hacked cells. The mitigation rate is computed over evaluable hacked cells, so it is not necessarily monotonic in the threshold because changing the threshold changes which hacked cells remain in the denominator.}
\label{FigThresholds}
\end{figure}

\subsection{What if the researcher strategically picks the configuration to preregister?}

In the earlier section, we analyzed a setting which implicitly assumed that the researcher will pick one hackable configuration at random from the current model and preregister it. However, the researcher could take different approaches to try and increase the chances of p-hacking under our protocol. We analyze two such approaches in this setting. 

The researcher could preregister the configuration that has the highest target-label rate on the current model. For instance, in the review classification task if the current model was Claude Opus 4.5, the researcher would choose the batched high temp configuration since it has the highest value. We call this the `highest value on latest model' strategy.

Alternatively, the researcher could execute the current and past models, and identify the configuration that has been the most hackable. For instance, in the review classification task if the current model was Claude Opus 4.5, then the researcher would preregister the counterfactual configuration. Even though the counterfactual configuration would not be hackable under the current model, it was hackable under three previous models, whereas all other configurations were hackable at most twice. We call this the `most often hacked so far' strategy.

We evaluate our protocol on such choices made by the researcher. We plot our results in Figure~\ref{FigMostHackablePrompt}. We see that the mitigation rates continue to be at least 60\% across all the settings, thereby showing that our protocol is robust to various possible strategies an opportunistic researcher could adopt. These results suggest that adversarial configuration selection can marginally reduce the protocol’s protection in some settings, but that preregistering for the next eligible model still substantially reduces successful carryover overall.

\definecolor{strategyred}{HTML}{AA0000}
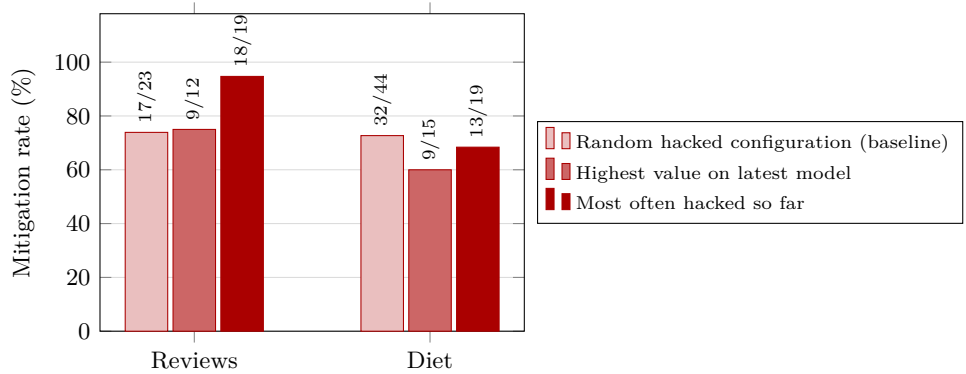
\begin{figure}[t]
    \centering
\begin{tikzpicture}
  \begin{axis}[
    ybar, bar width=16pt,
    scale only axis, width=0.34\textwidth, height=4.2cm,
    symbolic x coords={Reviews, Diet},
    xtick=data, enlarge x limits=0.4,
    ymin=0, ymax=118, ytick={0,20,...,100},
    ymajorgrids, grid style={gray!30, very thin},
    ylabel={Mitigation rate (\%)},
    tick label style={font=\small}, label style={font=\small},
    legend style={font=\scriptsize, at={(1.03,0.5)}, anchor=west,
                  legend columns=1},
    legend cell align=left,
    nodes near coords={\ratio},
    every node near coord/.append style={font=\scriptsize, color=black,
                                         rotate=90, anchor=west},
    visualization depends on={value \thisrow{ratio} \as \ratio},
  ]
    \addplot[fill=strategyred!25, draw=strategyred] table[x=expt, y=rate, row sep=\\] {
      expt rate ratio\\
      Reviews 73.9 17/23\\
      Diet 72.7 32/44\\
    };
    \addplot[fill=strategyred!60, draw=strategyred] table[x=expt, y=rate, row sep=\\] {
      expt rate ratio\\
      Reviews 75 9/12\\
      Diet 60 9/15\\
    };
    \addplot[fill=strategyred, draw=strategyred] table[x=expt, y=rate, row sep=\\] {
      expt rate ratio\\
      Reviews 94.7 18/19\\
      Diet 68.4 13/19\\
    };
    \legend{Random hacked configuration (baseline), Highest value on latest model, Most often hacked so far}
  \end{axis}
\end{tikzpicture}%
    \caption{Mitigation rate of LLM-based p-hacking under our protocol when the opportunistic researcher tries different strategies of choosing the configuration for preregistration. Mitigation remains substantial under both strategic choices.}
    \label{FigMostHackablePrompt}
\end{figure}

\subsection{What if the researcher chooses only one model class?} 

\definecolor{providergreen}{HTML}{004400}
\begin{figure}[ht]
\centering
\begin{tikzpicture}
  \begin{axis}[
    ybar, bar width=11pt,
    scale only axis, width=0.44\textwidth, height=4.2cm,
    symbolic x coords={Reviews, Diet},
    xtick=data, enlarge x limits=0.4,
    ymin=0, ymax=118, ytick={0,20,...,100},
    ymajorgrids, grid style={gray!30, very thin},
    ylabel={Mitigation rate (\%)},
    tick label style={font=\small}, label style={font=\small},
    legend style={font=\scriptsize, at={(1.03,0.5)}, anchor=west,
                  legend columns=1},
    legend cell align=left,
    nodes near coords={\ratio},
    every node near coord/.append style={font=\scriptsize, color=black,
                                         rotate=90, anchor=west},
    visualization depends on={value \thisrow{ratio} \as \ratio},
  ]
    \addplot[fill=gray!30, draw=gray] table[x=expt, y=rate, row sep=\\] {
      expt rate ratio\\
      Reviews 73.9 17/23\\
      Diet 72.7 32/44\\
    };
    \addplot[fill=providergreen!25, draw=providergreen] table[x=expt, y=rate, row sep=\\] {
      expt rate ratio\\
      Reviews 50 4/8\\
      Diet 66.7 6/9\\
    };
    \addplot[fill=providergreen!45, draw=providergreen] table[x=expt, y=rate, row sep=\\] {
      expt rate ratio\\
      Reviews 100 1/1\\
      Diet 50 1/2\\
    };
    \addplot[fill=providergreen!70, draw=providergreen] table[x=expt, y=rate, row sep=\\] {
      expt rate ratio\\
      Reviews 50 4/8\\
      Diet 64.3 9/14\\
    };
    \addplot[fill=providergreen, draw=providergreen] table[x=expt, y=rate, row sep=\\] {
      expt rate ratio\\
      Reviews 66.7 2/3\\
      Diet 100 3/3\\
    };
    \legend{All providers (baseline), Anthropic only, Google only, OpenAI only, xAI only}
  \end{axis}
\end{tikzpicture}%
\caption{Efficacy of our proposed protocol when considering just one model provider at a time. The protocol mitigates p-hackability by at least 50\% in all provider-restricted settings, but the single-provider results are heterogeneous and should be interpreted cautiously because several subsets have small denominators.}
\label{FigOneProvider}
\end{figure}
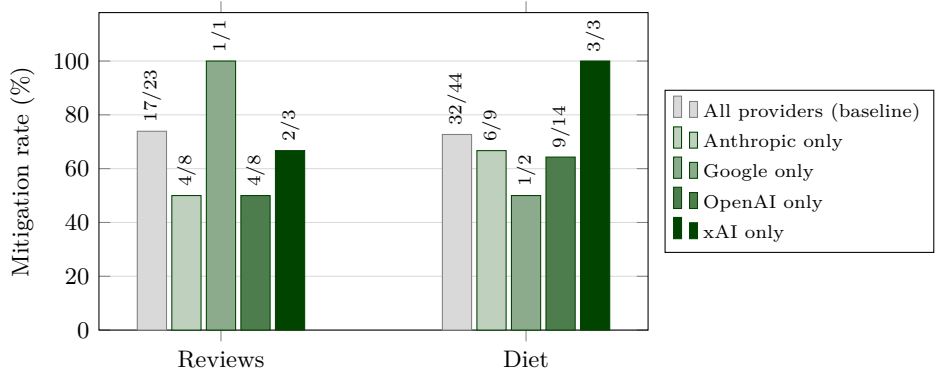

In our experiment, we preregistered and evaluated four model providers (OpenAI, Anthropic, Google, xAI). One may argue that models from the same provider may be more homogeneous, and a configuration that leads to p-hacking under a model from one provider is likely to be p-hackable under new models from the same provider. Then an opportunistic researcher may consider only one model provider for preregistration. Alternatively, the opportunistic researcher may try to predict which provider will release the next model among their preregistered set, and then preregister the configuration that results in p-hacking under that model provider. 

We investigate the efficacy of our proposed protocol under these strategies. In particular, we separately consider the different model providers. Assuming just one provider at a time, we evaluate the mitigation rate offered by our protocol.

We plot the results in Figure~\ref{FigOneProvider}. We see that the protocol mitigates p-hackability by at least 50\% for each individual provider. A caveat to note though is the small sample sizes for each model provider, due to the number of models released by each provider being much smaller than the aggregate.

Note that if the researcher preregisters just one model class, this will be transparent in the preregistration and against our recommendations. They may attempt to predict the next model release, but it is not clear how predictable that is.

\subsection{Are true effects detected with the next model?} 
A safeguard that suppresses hacking effects is only useful if it continues to retain detection power and does not suppress genuine effects. As an extreme example, a protocol that always reports ``no effect'' would achieve a 100\% mitigation rate. To complement our earlier mitigation analysis which measures how often hacks on null data fail to recur, we now analyze the protocol's sensitivity to genuine effects. We focus on the diet task here because its ground-truth calorie labels let us inject genuine effects of known strength by resampling the existing data, whereas the review task does not allow resampling to a known effect size.

For the diet task where ground truth labels vary, we resample the 100 items by their ground-truth label (whether Day 2 is truly lower in calories, established from the calorie totals) to inject a real effect of size \{55, 60, 65, ..., 95\}\%, replacing the true 50\% base rate. In this analysis, an effect is detected using the same thresholding rule: for a given model and configuration, we compute the fraction of valid corrected outputs labeled ``Day 2,'' and we detect an effect whenever that fraction reaches the preregistered threshold of 53\%. Recall from Section~\ref{sec:evalmeth} that this threshold corresponds to a one-sided binomial test at a known significance level, so thresholding on the ``Day 2'' rate is equivalent to a test on that rate.

We then define the \emph{detection rate} analogously to the mitigation rate, as the fraction of evaluable cells detected on one model that are also detected on the next released model. Formally, for any model $m$ and configuration $c$, we let $D(m,c) \in \{0,1\}$ take value $1$ if an effect (via the 53\% threshold) is detected under that model and configuration, and otherwise it takes the value $0$. With this notation in place, we then define the Detection Rate $
=
\frac{
\#\{(m,c): D(m,c)=1 \ \text{and}\ D(\mathrm{next}(m,c),c)=1\}
}{
\#\{(m,c): D(m,c)=1 \ \text{and}\ \mathrm{next}(m,c)\ \text{exists}\}
}$.

In addition, as a comparative baseline, we also replace the next model (as proposed in our protocol) with a re-measurement using the same model under data resampling -- this is the detection power in the absence of our protocol. 

\begin{figure}
\centering
\begin{tikzpicture}
  \begin{axis}[
      width=10cm, height=5.5cm,
      xlabel={Injected effect strength (true \% of items where Day~2 is lower)},
      ylabel={Detection rate},
      title={Diet experiment: detection of a true effect across model releases},
      xmin=54, xmax=95.5, ymin=0, ymax=1.02,
      xtick={55,60,65,70,75,80,85,90,95},
      grid=both, grid style={gray!20},
      legend pos=south east, legend cell align=left,
  ]
\addplot[draw=none, name path=protoU, forget plot] coordinates {(55,0.6659) (60,0.8246) (65,0.9208) (70,0.9660) (75,0.9842) (80,0.9926) (85,0.9954) (90,0.9977) (95,0.9985)};
\addplot[draw=none, name path=protoL, forget plot] coordinates {(55,0.5619) (60,0.7298) (65,0.8420) (70,0.9087) (75,0.9378) (80,0.9491) (85,0.9532) (90,0.9576) (95,0.9540)};
\addplot[blue!12, forget plot] fill between[of=protoU and protoL];
\addplot[draw=none, name path=baseU, forget plot] coordinates {(55,0.7090) (60,0.8445) (65,0.9321) (70,0.9722) (75,0.9878) (80,0.9938) (85,0.9971) (90,0.9985) (95,0.9990)};
\addplot[draw=none, name path=baseL, forget plot] coordinates {(55,0.6158) (60,0.7726) (65,0.8755) (70,0.9353) (75,0.9665) (80,0.9792) (85,0.9856) (90,0.9885) (95,0.9899)};
\addplot[black!12, forget plot] fill between[of=baseU and baseL];
\addplot[blue, mark=*, thick] coordinates {(55,0.6183) (60,0.7796) (65,0.8846) (70,0.9404) (75,0.9643) (80,0.9746) (85,0.9788) (90,0.9815) (95,0.9820)};
\addlegendentry{our protocol (confirm on the next model)}
\addplot[black!60, dashed, mark=square*, thick] coordinates {(55,0.6666) (60,0.8123) (65,0.9071) (70,0.9557) (75,0.9781) (80,0.9874) (85,0.9922) (90,0.9944) (95,0.9955)};
  \addlegendentry{baseline (re-measure with the same model)}
  \end{axis}
  \end{tikzpicture}
  
      \caption{Detection rate (power) of a genuine effect under our protocol on the diet experiment. We inject a true effect of a given strength (the proportion of items for which Day~2 is genuinely
  lower in calories). For each effect strength, datasets are generated by sampling items with replacement over 500 rounds. We then measure how often a detected (true) effect is re-detected by the next released model. As a baseline, we replace the next model with a re-measurement of the same model on resampled items. Shaded regions are bootstrapped 95\% confidence intervals. The confidence intervals overlap at every effect size, indicating that re-measuring with the next model (our protocol) costs no detectable power compared to re-measuring with the same model (not following our protocol).}
      \label{FigReplicationRate}
  \end{figure}
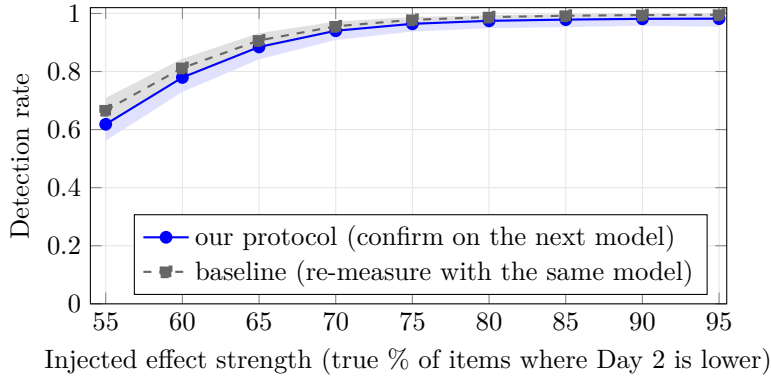

The results are shown in Figure~\ref{FigReplicationRate}, pooling across all models and configurations. We find that the detection rate under our protocol is very similar to that if our protocol was not used. Taken together with the results in the previous sections, we find that our protocol mitigates LLM-based p-hacking while retaining detection power.

\definecolor{strategyred}{HTML}{AA0000}

\subsection{Time between model releases}
A downside of our protocol is that the researcher has to wait for the next model release after preregistration. For our preregistered set of models, in Figure~\ref{fig:release-timeline} we plot the release dates and the gap between releases. If a researcher were to preregister on a day chosen uniformly at random between August 7, 2025 and May 27, 2026 -- the first and last releases in our study -- then their \textbf{mean and median waiting time for the next model would be 14.5 days and 10.3 days respectively}. Finally, observe that the largest gap between model releases occurs during the winter break of December 2025 - January 2026.

\definecolor{anthropic}{HTML}{D97757}
\definecolor{openai}{HTML}{10A37F}
\definecolor{googleblue}{HTML}{4285F4}
\definecolor{xai}{HTML}{444444}

\begin{figure*}[t]
\centering
\begin{tikzpicture}
\newcommand{\rel}[5]{%
    \node[circle, fill=#4, inner sep=2.0pt] (n) at (axis cs:#2,#3) {};
    \node[#1=2.5pt of n, font=\tiny, inner sep=0pt, text=#4!50!black] {#5};
}
\begin{groupplot}[
    group style={
        group size=1 by 2,
        x descriptions at=edge bottom,
        vertical sep=4pt,
    },
    width=\textwidth,
    date coordinates in=x,
    xmin=2025-08-01, xmax=2026-06-14,
    xtick={2025-08-01,2025-09-01,2025-10-01,2025-11-01,2025-12-01,2026-01-01,
           2026-02-01,2026-03-01,2026-04-01,2026-05-01,2026-06-01},
    xticklabels={},
    xmajorgrids,
    grid style={black!12},
    tick label style={font=\footnotesize},
    label style={font=\footnotesize},
    axis line style={black!60},
    every tick/.style={black!60},
    clip=false,
]

\nextgroupplot[
    height=4.6cm,
    ymin=0.4, ymax=4.6,
    ytick={1,2,3,4},
    yticklabels={xAI, Google, OpenAI, Anthropic},
]
\rel{above}{2025-09-29}{4}{anthropic}{Claude Sonnet 4.5}
\rel{above}{2025-11-24}{4}{anthropic}{Claude Opus 4.5}
\rel{above}{2026-02-04}{4}{anthropic}{Claude Opus 4.6}
\rel{below}{2026-02-17}{4}{anthropic}{Claude Sonnet 4.6}
\rel{above}{2026-04-16}{4}{anthropic}{Claude Opus 4.7}
\rel{above}{2026-05-27}{4}{anthropic}{Claude Opus 4.8}
\rel{above}{2025-08-07}{3}{openai}{GPT 5 Chat}
\rel{above}{2025-11-13}{3}{openai}{GPT 5.1 Chat}
\rel{below}{2025-12-10}{3}{openai}{GPT 5.2 Chat}
\rel{above}{2026-03-03}{3}{openai}{GPT 5.3 Chat}
\rel{below}{2026-03-05}{3}{openai}{GPT 5.4}
\rel{below}{2026-04-24}{3}{openai}{GPT 5.5}
\rel{above}{2025-11-18}{2}{googleblue}{Gemini 3 Pro Preview}
\rel{below}{2025-12-17}{2}{googleblue}{Gemini 3 Flash Preview}
\rel{above}{2026-02-19}{2}{googleblue}{Gemini 3.1 Pro Preview}
\rel{above}{2026-05-19}{2}{googleblue}{Gemini 3.5 Flash}
\rel{above}{2025-09-19}{1}{xai}{Grok 4 Fast}
\rel{above}{2025-11-19}{1}{xai}{Grok 4.1 Fast}
\rel{above}{2026-03-31}{1}{xai}{Grok 4.20}
\rel{above}{2026-04-30}{1}{xai}{Grok 4.3}

\nextgroupplot[
    height=3.2cm,
    ymin=0, ymax=54,
    ylabel={Days since\\previous release},
    ylabel style={align=center},
    ymajorgrids,
    every axis plot/.append style={
        ycomb, mark=*, mark size=1.6pt, line width=2.4pt},
    extra x ticks={2025-08-15,2025-09-15,2025-10-15,2025-11-15,2025-12-15,
                   2026-01-15,2026-02-15,2026-03-15,2026-04-15,2026-05-15},
    extra x tick labels={Aug~'25,Sep~'25,Oct~'25,Nov~'25,Dec~'25,
                         Jan~'26,Feb~'26,Mar~'26,Apr~'26,May~'26},
    extra x tick style={grid=none, tick style={draw=none}},
]
\addplot[anthropic] coordinates {
    (2025-09-29,10) (2025-11-24,5) (2026-02-04,49) (2026-02-17,13) (2026-04-16,16) (2026-05-27,8)};
\addplot[openai] coordinates {
    (2025-11-13,45) (2025-12-10,16) (2026-03-03,12) (2026-03-05,2) (2026-04-24,8)};
\addplot[googleblue] coordinates {
    (2025-11-18,5) (2025-12-17,7) (2026-02-19,2) (2026-05-19,19)};
\addplot[xai] coordinates {
    (2025-09-19,43) (2025-11-19,1) (2026-03-31,26) (2026-04-30,6)};
\end{groupplot}
\end{tikzpicture}
\caption{Release timeline of the evaluated models (top) and number of days
elapsed between consecutive releases across all providers (bottom). Colors
denote the provider, and the vertical gridlines indicate the first day of each month.}
\label{fig:release-timeline}
\end{figure*}
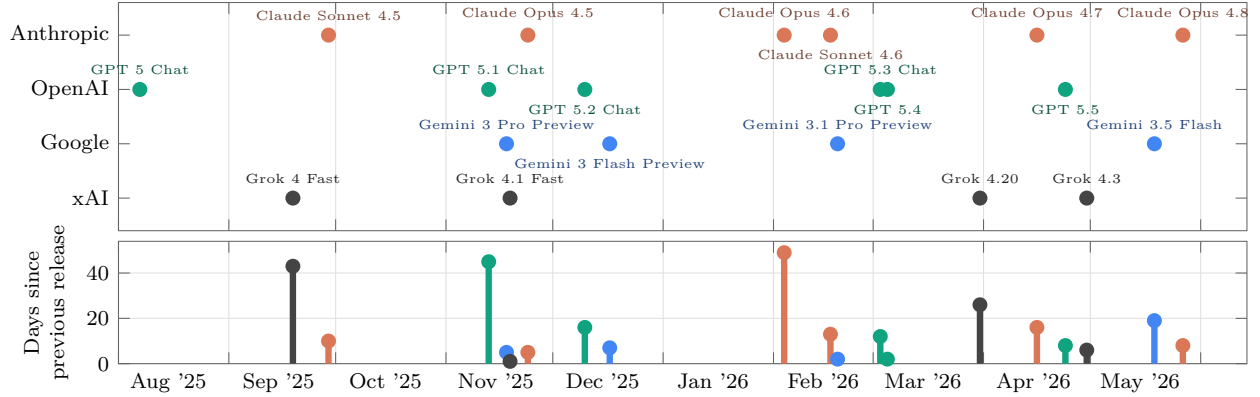

\section{Related work}

\paragraph{Heuristic validation.} A growing body of work studies whether and how LLM-generated data can support valid statistical inferences. The first line of work establishes validity through heuristic validation~\citep{hullman2026human}: researchers demonstrate that LLM responses resemble ground-truth responses on an overlapping set (matching the direction, significance, size of effects, or distributional statistics) and treat the two as interchangeable for statistical tests in related settings where human ground truth is unavailable \citep{argyle2023out, kozlowski2024simulating}. P-values are then interpreted just as they are in traditional analyses. To make LLM data resemble ground-truth data, existing repair strategies optimize the LLMs before statistical estimation, via prompt engineering, persona design, response formatting, and fine-tuning \citep{binz2025foundation, kolluri2025finetuning, suh2025language,krsteski2025valid}. But \textbf{this does not solve p-hacking}: every repair strategy is \emph{itself} a researcher degree of freedom, and a researcher can tune these choices toward statistical significance of the final test. The same flexibility that aligns LLM instruments with ground-truth data is what makes them hackable. Unlike these existing strategies which aim to argue for the validity of a given LLM instrument, here, we safeguard the inferential process itself through a preregistration mechanism.

\paragraph{Debiasing.} The second line of work combines a small ground-truth sample with a larger LLM-based sample to recover asymptotically valid statistical estimation. Prediction-powered inference (PPI) \citep{angelopoulos2023prediction}, its application to mixed-subjects designs~\citep{broska2025mixed}, and active versions~\citep{gligoric2025can}, design-based supervised learning (DSL)~\citep{egami2023using}, plug-in bias correction~\citep{ludwig2025large}, and doubly-robust extensions~\citep{guerdan2025doubly} all estimate a rectifier from overlapping data and use it to debias statistical estimation. These methods produce statistically valid estimates, but, \textbf{they do not safeguard against p-hacking}. Moreover, these methods restore valid inference but trade off Type I vs. Type II errors~\citep{baumann2025}, i.e., producing overly conservative estimates in finite samples, and missing true effects. \textbf{In contrast, our protocol retains power and continues to detect true effects.} \citeauthor{baumann2025} studied these methods in the context of LLM hacking~\citep{baumann2025}, but, a key distinction compared to such existing analyses is that we do not rely on human annotations and statistical analyses based on them as ground truth; instead, we evaluate on settings where we know effects do not exist.

\paragraph{Robustness tests.} Lastly, a third practice reports results across many specifications, and treats stability across them as evidence of validity (in the spirit of multiverse and specification-curve analysis; \citealp{ye2026stop,steegen2016increasing,simonsohn2020specification}). However, \textbf{this is inadequate as a p-hacking safeguard}. The choice of which robustness checks to run and report is itself an unconstrained degree of freedom, so selective reporting reproduces the original issue~\citep{gelman2013garden}. Stability across a hand-picked set of specifications places no bound on the error over the full space of forking paths, and provides no formalized and reproducible procedure. Checklists help to make these disclosures more principled~\citep{goldberg2024checklist,llmchecklist}, but they have been shown to have limited effectiveness, since they rely solely on self-report~\citep{wong2020beyond}. These shortcomings motivate extending the preregistration paradigm which can empower researchers to be transparent and commit to an LLM analysis pipeline before the new model is released and available to hack against.

\section{Discussion}\label{sec:diss}

We propose extending preregistration to LLM-based research to mitigate p-hacking: finalize the procedure on current models, commit to it and to a set of eligible future models, then run the confirmatory analysis on the first eligible model released afterward. In our experiments, we find that this approach  can mitigate the amount of p-hacking.

The need for such safeguards is growing with the increasing prevalence of autonomous AI scientists. These systems conduct end-to-end research~\citep{luEndtoendAutomationAI2026,yamada2025ai,schmidgallAgentLaboratoryUsing2025} with little or no human intervention. However, these systems have been found to engage in p-hacking~\citep{luo2025}. Our protocol mitigates this directly, and we propose integrating it into autonomous research frameworks: whenever such a system runs an LLM-based analysis, it should commit to the procedure and eligible-model set before the confirmatory model is available.

A possible objection is that a researcher could privately preregister many configurations and then report only the one that happens to remain hackable on the next model. This concern is not unique to our setting---it applies to preregistration in general, and it has been studied previously~\citep{simmons2021pre}. In practice, registries mitigate it through both technical and social mechanisms: for instance, AsPredicted flags and blocks near-duplicate submissions, while platforms such as OSF require preregistrations to become public after an embargo~\citep{simmons2021pre}. The same safeguards apply to LLM preregistrations, and future efforts to build a public registry of all preregistered LLM-based analyses would further constrain selective reporting in our setting. 

Another possible objection is that one could obtain similar protection simply by delaying the study and using a newer model. But delaying without committing leaves the researcher free to search configurations against the new model and report the one that hacks; since a single hackable configuration suffices, a lower base rate does not offer much protection (the researcher could search well beyond the eleven configurations we tested). Indeed, four of the six models released after our preregistration still admitted at least one hacking configuration. The protocol's protection comes from removing this search through commitment. Even where a model is genuinely robust, committing in advance still gives readers a verifiable signal that the result was not searched for.

\paragraph{Limitations.} Importantly, our protocol is not without its limitations. It reduces the chance that a p-hack carries over, but does not eliminate it. As our results show, some configurations do propagate to the next model. Even then, the researcher still benefits by following the procedure because the analysis is committed in advance and is transparent, which is itself a basis for trust. The protocol shifts LLM-based research from an open-ended search over forking paths to a commitment that authors can demonstrate at submission time. 

Moreover, the scope of our empirical evaluation is focused on four model families and two tasks, both of which are binary classification problems, and each draws on 100 labeled items from two datasets, analyzed using thresholds; whether the protocol's effectiveness extends to other tasks (such as multi-class labeling, free-form generation, regression-style outputs), other types of statistical analyses, model families, or other corpora remains to be evaluated in future work.

Finally, although we frame the threat as p-hacking, we operationalize a hack with a fixed threshold on the target-label rate rather than computing p-values directly. This is partly by necessity as the review task's null of zero AI prevalence is a boundary null at which a p-value is degenerate. Where a non-degenerate null exists, as in the dietary task, the threshold already corresponds to a one-sided binomial test at a known significance level. A fuller instantiation that computes p-values across conventional significance levels and incorporates multiple-testing corrections remains future work.

\paragraph{Conclusion.} We envision that venues could encourage this form of preregistration for any work leveraging LLM-generated data, annotations, or using LLM-as-a-judge. Many venues already similarly encourage sharing code and data, preregistering human-subject experiments, or completing reproducibility and ethical checklists. Importantly, such encouragement would work with rather than against authors' existing incentives: as models improve, researchers may already want to adopt the strongest available model for better reasoning, so committing in advance to the next eligible release asks little beyond their existing practice. The cost is the short wait for the next release, a median of approximately ten days in our study---a modest price for demonstrably trustworthy analyses.

\section*{Acknowledgments} 
The work of NBS was supported by grants NSF 1942124 and ONR N000142512346.

\bibliographystyle{plainnat}
\bibliography{references}

\section*{Appendix: Configurations}

\begin{itemize}
    \item \textbf{Zero-Shot}: This is the primary comparison prompt. It presents the task with minimal additional framing: for the review classification task, the model classified each review as \texttt{AI} or \texttt{Human}; for the dietary comparison task, the model selected whether \texttt{Day 1} or \texttt{Day 2} had the lower estimated calorie intake.

    \item \textbf{Counterfactual}: This prompt adds a brief counterfactual instruction before the final classification. In the review classification task, the model is asked to consider evidence that the review could be AI-generated. In the dietary comparison task, the model is asked to consider evidence that \texttt{Day 2} could have lower calorie intake.

    \item \textbf{Instructional}: This role-framed version provides more detailed task instructions and decision criteria. For the review classification task, the prompt emphasizes using the review's writing style and content. For the dietary comparison task, it emphasizes comparing foods, portions, and calorie-relevant details across the two diary days.

    \item \textbf{Directional}: This prompt introduces subtle directional language that could influence the model's interpretation. For reviews, polished or generic wording was presented as potential evidence of AI-generation. For diet diaries, \texttt{Day 2} is framed as a follow-up day where healthier substitutions could indicate lower calorie intake.

    \item \textbf{Few-Shot}: This condition includes a small number of labeled examples before the target item. The purpose is to evaluate whether in-context examples affect the model's classification behavior on the review classification and dietary comparison tasks.

    \item \textbf{Framed + Imputation Rules}: This prompt combines stronger contextual framing with explicit rules for handling ambiguous cases. For the review classification task, the prompt describes the AI-detection approach as promising and provides guidance for weighing generic phrasing and unclear evidence. For the dietary comparison task, the prompt describes the intervention as promising for reducing caloric intake and includes rules for interpreting missing portions, vague descriptions, and assumptions.

    \item \textbf{Low Top-p}: This condition uses the same single-item prompt text as \textbf{Zero-Shot} but changes the decoding settings to \texttt{top\_p = 0.3} and \texttt{temperature = 0.0}. This tests whether restricting the sampling distribution affects classifications.

    \item \textbf{High Temp}: This condition uses the same single-item prompt text as \textbf{Zero-Shot} but changes the decoding settings to  \texttt{top\_p = 1.0} and \texttt{temperature = 2.0}. This tests whether increased sampling randomness affects classification outcomes.

    \item \textbf{Batched Zero-Shot}: This condition presents ten items in a single prompt instead of one item at a time. The model is required to return a structured JSON response containing one prediction for each review or diet pair.

    \item \textbf{Batched Low Top-p}: This condition uses the same batch-of-ten JSON format as \textbf{Batched Zero-Shot}, but with \texttt{top\_p = 0.3} and \texttt{temperature = 0.0}. This tests whether restricting the sampling distribution affects predictions in the batch setting.
    
    \item \textbf{Batched High Temp}: This condition uses the same batch-of-ten JSON format as \textbf{Batched Zero-Shot}, but with  \texttt{top\_p = 1.0} and \texttt{temperature = 2.0}. This tests whether increased sampling randomness has different effects in the batch setting.

\end{itemize}

\end{document}